**Title Page.**

Aickelin U and Li J (2007): 'An Estimation of Distribution Algorithm for Nurse Scheduling', Annals of Operations Research, in print, doi: 10.1007/s10479-007-0214-0, pp TBA.

**Paper title:**

An Estimation of Distribution Algorithm for Nurse Scheduling

**Authors (alphabetical order):**

UWE AICKELIN
uxa@cs.nott.ac.uk
School of Computer Science and Information Technology
University of Nottingham
Nottingham, NG8 1BB
UK

JINGPENG LI
jpl@cs.nott.ac.uk
School of Computer Science and Information Technology
University of Nottingham
Nottingham, NG8 1BB
UK

**Contact author:**

Dr JINGPENG LI
School of Computer Science & Information Technology
Jubilee Campus
University of Nottingham
Nottingham, NG8 1BB
UK
Email: jpl@cs.nott.ac.uk
Tel: +44(0) 115 8466527
Fax: +44(0) 115 9514251

**Abstract.**

Schedules can be built in a similar way to a human scheduler by using a set of rules that involve domain knowledge. This paper presents an Estimation of Distribution Algorithm (EDA) for the nurse scheduling problem, which involves choosing a suitable scheduling rule from a set for the assignment of each nurse. Unlike previous work that used Genetic Algorithms (GAs) to implement implicit learning, the learning in the proposed algorithm is explicit, i.e. we identify and mix building blocks directly. The EDA is applied to implement such explicit learning by building a Bayesian network of the joint distribution of solutions. The conditional probability of each variable in the network is computed according to an initial set of promising solutions. Subsequently, each new instance for each variable is generated by using the corresponding conditional probabilities, until all variables have been generated, i.e. in our case, a new rule string has been obtained. Another set of rule strings will be generated in this way, some of which will replace previous strings based on fitness selection. If stopping conditions are not met, the conditional probabilities for all nodes in the Bayesian network are updated again using the current set of promising rule strings. Computational results from 52 real data instances demonstrate the success of this approach. It is also suggested that the learning mechanism in the proposed approach might be suitable for other scheduling problems.

**Text.**

Nurse scheduling problems have been widely studied in recent years. They are generally NP-hard combinatorial problems, and a lot of research has therefore been done to solve these problems heuristically or meta-heuristically (e.g., Aickelin and Dowsland (2000, 2004); Bellanti et al. (2004); Burke et al. (1999, 2001); Dowsland and Thompson (2000); Isken and Hancock (1991); Jan et al. (2000); Jaszkiewicz (1997)). A comprehensive review of the approaches to nurse rostering can be found in Burke et al. (2004) and Cheang et al. (2003). Genetic Algorithms (GAs) (Holland (1975), Goldberg (1989)) form an important class of meta-heuristics. They are search techniques based on selection and recombination of promising solutions, and have achieved great success in solving difficult scheduling problems in recent years. However, the performance of GAs depends on the choice of the genetic operators (such as selection, crossover, mutation, probabilities of crossover and mutation, population size, number of generations etc.), and problem specific interactions among the variables (genes) are seldom considered (Aickelin (2002)). Therefore, crossover recombination and evolution sometimes results in premature convergence. To avoid the disruption of partial solutions that consist of important building blocks, the crossover recombination processes can be replaced by generating new solutions according to the probability distribution of all promising solutions in the previous generation. This new approach is called Estimation of Distribution Algorithm (EDA) (Larranaga and Lozano (2001)). In this paper, we propose such an EDA for nurse scheduling.

Nurse scheduling is usually highly constrained, making it extremely difficult for most local search algorithms to find feasible solutions, let alone optimal ones. In our nurse scheduling problem, the number of nurses is fixed (up to 30), and the target is to create a weekly schedule by assigning each nurse one out of up to 411 shift patterns in the most efficient way. The proposed EDA approach achieves this by choosing a suitable rule, from a set containing a number of available rules, for each nurse. A solution is therefore represented as a rule string, or a sequence of rules corresponding to nurses from the first one to the last.

To evolve the rule strings, a Bayesian network (Pearl (1998)) is used in the proposed EDA to solve the nurse scheduling problem. A Bayesian network is a directed acyclic graph with each node corresponding to one variable, and each variable corresponding to the individual rule by which a schedule will be constructed step by step. The causal relationship between two variables is represented by a directed edge between the two corresponding nodes (see figures 1 and 2).

The EDA learns to identify good partial solutions and to complete them by building a Bayesian network of the joint distribution of solutions (Pelikan et al. (1999), Pelikan and Goldberg (2000)). The conditional probabilities are computed according to an initial set of promising solutions. Subsequently, each new instance for each node is generated by using the corresponding conditional probabilities, until values for all nodes have been generated, i.e. a new rule string has been generated. Thus, another set of rule strings can be generated in this way, some of which will replace previous strings based on roulette-wheel fitness selection. If stopping conditions are not met, the conditional probabilities for all nodes in the Bayesian network are updated again using the current set of rule strings. The algorithm thereby tries to implicitly identify and mix promising building blocks.

In EDAs, the structure of the Bayesian network can be either fixed (Pelikan et al. (1999)) or variable (Mühlenbein and Mahnig (1999)). In our nurse scheduling model, we use a fixed network structure because all variables are fully observed. In essence, our network is a fixed, nurse-size vector of rules and the goal of learning is to find the rule values that maximize the likelihood of the training data. Thus, our learning amounts to 'counting' based on a multinomial distribution.

The rest of this paper is organized as follows: Section 1 gives an overview of the nurse scheduling problem and section 2 introduces the general concepts about graphical models and Bayesian networks. Section 3 discuses the proposed EDA in detail. Section 4 presents computational results using 52 data instances gathered from a UK hospital, and suggests several directions for further research. Concluding remarks are in section 5.

## 1 The nurse scheduling problem

## 1.1 General problem

Our nurse scheduling problem is to create weekly schedules for wards of nurses by assigning one of a number of possible shift patterns to each nurse. These schedules have to satisfy working contracts and meet the demand for a given number of nurses of different grades on each shift, while being seen to be fair by the staff concerned. The latter objective is achieved by meeting as many of the nurses' requests as possible and considering historical information to ensure that unsatisfied requests and unpopular shifts are evenly distributed.

The problem is complicated by the fact that higher qualified nurses can substitute less qualified nurses but not vice versa. Thus scheduling the different grades independently is not possible. Furthermore, the problem has a special day-night structure as most of the nurses are contracted to work either days or nights in one week but not both. However, due to working contracts, the number of days worked is not usually the same as the number of nights. Therefore, it becomes important to schedule the 'correct' nurses onto days and nights respectively. The latter two characteristics make this problem challenging for any local search algorithm, because finding and maintaining feasible solutions is extremely difficult.

The numbers of days or nights to be worked by each nurse defines the set of feasible weekly work patterns for that nurse. These will be referred to as shift patterns or shift pattern vectors in the following. For each nurse $i$ and each shift pattern $j$ all the information concerning the desirability of the pattern for this nurse is captured in a single numeric preference cost $p_{ij}$. These costs were determined in close consultation with the hospital and are a weighted sum of the following factors: basic shift-pattern cost, general day/night preferences, specific requests, continuity problems, number of successive working day, rotating nights/weekends and other working history information. Patterns that violate mandatory contractual requirements are marked as infeasible for a particular nurse and week by giving them a suitably high $p_{ij}$ value.

## 1.2 Integer programming

The problem can be formulated as an integer linear program as follows.

Indices:

$i = 1...n$ nurse index;

$j = 1...m$ shift pattern index;

$k = 1...14$ day and night index (1...7 are days and 8...14 are nights);

$s = 1...p$ grade index.

Decision variables:

$$x_{ij} = \begin{cases} 1, & \text{nurse } i \text{ works shift pattern } j \\ 0, & \text{else} \end{cases}.$$

Parameters:

$m$ = Number of shift patterns;

$n$ = Number of nurses;

$p$ = Number of grades;

$$a_{jk} = \begin{cases} 1, & \text{shift pattern } j \text{ covers day/night } k \\ 0, & \text{else} \end{cases};$$

$$q_{is} = \begin{cases} 1, & \text{nurse } i \text{ is of grade } s \text{ or higher} \\ 0, & \text{else} \end{cases};$$

$p_{ij}$ = Preference cost of nurse $i$ working shift pattern $j$;

$F(i)$ = Set of feasible shift patterns for nurse $i$;

$D_i$ = Working shifts per week of nurse $i$ if day shifts are worked;

$N_i$ = Working shifts per week of nurse $i$ if night shifts are worked; $B_i$ = Working shifts per week of nurse $i$ if both day and night shifts are worked {for special nurses};

$R_{ks}$ = Demand of nurses with grade $s$ on day/night $k$;

Target function:

Minimize total preference cost of all nurses, denoted as

$$\sum_{i=1}^{n} \sum_{j \in F(i)}^{m} p_{ij} x_{ij} \to \min! \quad (1)$$

Subject to:

Every nurse works exactly one feasible shift pattern:

$$\sum_{j \in F(i)} x_{ij} = 1, \forall i; \quad (2)$$

The shift pattern corresponds to the number of weekly working shifts of the nurse:

$$F(i) = \begin{cases} \sum_{k=1}^{7} a_{jk} = D_i, & \forall j \in \text{day shifts} \\ \sum_{k=8}^{14} a_{jk} = N_i, & \forall j \in \text{night shifts} \\ \sum_{k=1}^{14} a_{jk} = B_i, & \forall j \in \text{combined shifts} \end{cases}, \forall i. \quad (3)$$

The demand for nurses is fulfilled for every grade on every day and night:

$$\sum_{j \in F(i)} \sum_{i=1}^{n} q_{is} a_{jk} x_{ij} \geq R_{ks}, \forall k, s \quad (4)$$

(1) gives the objective and constraint sets (2) and (3) ensure that every nurse works exactly one shift pattern from his/her feasible set, and constraint set (4) ensures that the demand for nurses is covered for every grade on every day and night. Note that the definition of $q_{is}$ is such that higher graded nurses can substitute those at lower grades if necessary.

Typical problem dimensions are 30 nurses of three grades and 400 shift patterns. Thus, the Integer Programming formulation has about 12000 binary variables and 100 constraints. This is a moderately sized problem. However, some problem cases remain unsolved after overnight computation using a commercial integer programming software package called XPRESS MP (Dowsland and Thompson (2000)).

## 2 Graphical models and Bayesian networks

In this section, we introduce concepts from graphical models in general and Bayesian networks in particular. Section 3 will then explain how we applied these concepts to our nurse scheduling problem.

Graphical models are graphs in which nodes represent random variables, and edges represent conditional dependence assumptions (Edwards (2000)). They have been applied to many multivariate probabilistic systems in fields such as statistics, systems engineering, information theory and pattern recognition. In particular, they are playing an increasingly important role in the design and analysis of machine learning algorithms.

As described by Jordon (1999), graphical models are a marriage between probability theory and graph theory. They provide a natural tool for dealing with uncertainty and complexity that occur throughout applied mathematics and engineering. In a graphical model, the fundamental notion of modularity is used to build a complex system by combining simpler parts. Probability theory provides the glue to combine the parts, ensuring that the whole system is consistent, and providing ways to interface models to data. Graph theory provides an intuitively appealing interface by which humans can model highly interacting sets of variables, and a data structure that leads itself naturally to the design of general-purpose algorithms.

There are two main kinds of graphical models: undirected and directed. Undirected graphical models are more popular with the physics and vision communities. Directed graphical models, also called Bayesian networks, are more popular with the artificial intelligence and machine learning communities. Bayesian networks are often used to model multinomial data with both discrete and continuous variables by encoding the relationship between the variables contained in the modelled data, which represents the structure of a problem.

Moreover, Bayesian networks can be used to generate new data instances or variables instances with similar properties as those of given data. Each node in the network corresponds to one variable, and each variable corresponds to one position in the strings representing the solutions. The relationship between two variables is represented by a directed edge between the two corresponding nodes.

Any complete probabilistic model of a domain must represent the joint distribution, the probability of every possible event as defined by the values of all the variables. The number of such events is exponential: if we have *n* binary nodes, the full joint needs $O(2^n)$ parameters. To achieve compactness, Bayesian networks factor the joint distribution into local conditional distributions for each variable given its parents. For example, by the chain rule of probability, the joint probability of all nodes in Figure 1 is:

$$P(x_1, x_2, x_3, x_4) = P(x_1) \times P(x_2|x_1) \times P(x_3|x_1, x_2) \times P(x_4|x_1, x_2, x_3) \tag{5}$$

However, using conditional independence relationships, we can rewrite this as

$$P(x_1, x_2, x_3, x_4) = P(x_1) \times P(x_2|x_1) \times P(x_3|x_1) \times P(x_4|x_2, x_3) \tag{6}$$

where the third term can be simplified because $x_3$ is independent of $x_2$ given its parents $x_1$, and the last term because $x_4$ is independent of $x_1$ given its parents $x_2$ and $x_3$. According to the above example, we can see that the conditional independence relationship allows us to represent the joint probability more compactly.

**Figure 1. Example of dependency graph with four nodes.**

Mathematically, an acyclic Bayesian network encodes a full joint probability distribution by the product

$$P(x_1, \ldots, x_n) = \prod_{i=1}^{n} P(x_i \mid pa(X_i)), \tag{7}$$

where $x_i$ denotes some values of the variable $X_i$, $pa(X_i)$ denotes a set of values for parents of $X_i$ in the network (the set of nodes from which there exists an individual edge to $X_i$), and $P(x_i \mid pa(X_i))$ denotes the conditional probability of $X_i$ conditioned on variables $pa(X_i)$, which can be used to generate new instances.

## 3 An EDA for nurse scheduling

This section discusses the proposed EDA for the nurse scheduling problem, including the construction of a Bayesian network, learning based on the Bayesian network and the four building rules used.

### 3.1 The construction of a Bayesian network

In our nurse scheduling problem, the number of nurses is fixed (up to 30), and the goal is to create weekly schedules by assigning each nurse one shift pattern in the most efficient way. The proposed approach achieves this by using one suitable rule, from a rule set that contains a number of available rules, for each nurse's assignment. Thus, a potential solution is represented as a rule string, or a sequence of rules corresponding to nurses from the first one to the last one individually.

We chose this approach, as the longer-term aim of our research is to model the explicit learning of a human scheduler. Human schedulers can provide high quality solutions, but the task is tedious and often requires a large amount of time. Typically, they construct schedules based on rules learnt during scheduling. Due to human limitations, these rules are typically simple. Hence, our rules will be relatively simple, too. Nevertheless, human generated schedules are of high quality due to the ability of the scheduler to switch between the rules, based on the state of the current solution. We envisage the proposed EDA to perform this role.

**Figure 2. A Bayesian network for nurse scheduling.**

Figure 2 is the Bayesian network constructed for the nurse scheduling problem, which is a hierarchical and acyclic directed graph representing the solution structure of the problem.

The node $N_{ij}(i \in \{1,2,...,m\}; j \in \{1,2,...,n\})$ in the network denotes that nurse *i* is assigned using rule *j*, where *m* is the number of nurses to be scheduled and *n* is the number of rules to be used in the building process. The directed edge from node $N_{ij}$ to node $N_{i+1,j'}$ denotes a causal relationship of "$N_{ij}$ causing $N_{i+1,j'}$". In our particular implementation, an edge denotes a building block (i.e. rule sub-string) for nurse *i* where the previous rule is *j* and the current rule is *j'*. In this network, a possible solution (a complete rule string) is represented as a directed path from nurse 1 to nurse *m* connecting *m* nodes.

**3.2 Learning based on the Bayesian network**

According to whether the structure (topology) of the model is known or unknown, and whether all variables are fully observed or some of them are hidden, there are four kinds of learning (Heckerman (1998)). The learning in our proposed EDA belongs to the category of "known structure and full observation," and the goal is to find the variable values of all nodes $N_{ij}$ that maximize the likelihood that the training data contains *T* independent cases.

In our EDA, learning amounts to counting and hence we use the symbol '#' meaning 'the number of' in the following equations. It calculates the conditional probabilities of each possible value for each node given all possible values of its parents. For example, for node $N_{i+1,j'}$ with a parent node $N_{ij}$, its conditional probability is

$$P(N_{i+1,j'} | N_{ij}) = \frac{P(N_{i+1,j'}, N_{ij})}{P(N_{ij})} = \frac{\#(N_{i+1,j'} = true, N_{ij} = true)}{\#(N_{i+1,j'} = true, N_{ij} = true) + \#(N_{i+1,j'} = false, N_{ij} = true)}. \tag{8}$$

Note that nodes $N_{1j}$ have no parents. In this circumstance, their probabilities are computed as

$$P(N_{1j}) = \frac{\#(N_{1j} = true)}{\#(N_{1j} = true) + \#(N_{1j} = false)} = \frac{\#(N_{1j} = true)}{T}. \tag{9}$$

To make it easier to understand how these probabilities are computed, let us consider a simple example by using three rules to schedule three nurses (shown in Figure 3). The scheduling process is repeated 50 times, and each time, rules are randomly used to get a solution, no matter if it is feasible or not. The Arabic numeral adjacent to each edge is the total number of times that this edge has been used in the 50 runs. For instance, if a solution is obtained by using rule 2 to schedule nurse 1, rule 3 to nurse 2 and rule 1 to nurse 3, then there exists a path of "$N_{12} \rightarrow N_{23} \rightarrow N_{31}$", and the count of edge "$N_{12} \rightarrow N_{23}$" and edge "$N_{23} \rightarrow N_{31}$" are increased by one respectively.

**Figure 3. A dummy problem with three nurses and three rules.**

Therefore, we can calculate the probabilities of each node at different states according to the above count. For the nodes that have no parents, their probabilities are computed as:

$$P(N_{11}) = \frac{10+2+3}{(10+2+3)+(5+11+4)+(7+5+3)} = \frac{15}{50}, \; P(N_{12}) = \frac{5+11+4}{50} = \frac{20}{50}, \; P(N_{13}) = \frac{15}{50}. \tag{10}$$

For all other nodes (with parents), the conditional probabilities are:

$$P(N_{21}|N_{11}) = \frac{10}{10+2+3} = \frac{10}{15}, \; P(N_{22}|N_{11}) = \frac{2}{10+2+3} = \frac{2}{15}, \; P(N_{23}|N_{11}) = \frac{3}{10+2+3} = \frac{3}{15},$$

$$P(N_{21}|N_{12}) = \frac{5}{5+11+4} = \frac{5}{20}, \; P(N_{22}|N_{12}) = \frac{10}{5+11+4} = \frac{10}{20}, \; P(N_{23}|N_{12}) = \frac{4}{5+11+4} = \frac{4}{20},$$

$$P(N_{21}|N_{13}) = \frac{7}{7+5+3} = \frac{7}{15}, \; P(N_{22}|N_{13}) = \frac{5}{7+5+3} = \frac{5}{15}, \; P(N_{23}|N_{13}) = \frac{3}{7+5+3} = \frac{3}{15},$$

$$P(N_{31}|N_{21}) = \frac{7}{7+9+3} = \frac{7}{19}, \; P(N_{32}|N_{21}) = \frac{9}{7+9+3} = \frac{9}{19}, \; P(N_{33}|N_{21}) = \frac{3}{7+9+3} = \frac{3}{19},$$

$$P(N_{31}|N_{22}) = \frac{11}{11+1+5} = \frac{11}{17}, \; P(N_{32}|N_{22}) = \frac{1}{11+1+5} = \frac{1}{17}, \; P(N_{33}|N_{22}) = \frac{5}{11+1+5} = \frac{5}{17},$$

$$P(N_{31}|N_{23}) = \frac{10}{10+4+0} = \frac{10}{14}, \; P(N_{32}|N_{23}) = \frac{4}{10+4+0} = \frac{4}{14}, \; P(N_{33}|N_{23}) = \frac{0}{10+4+0} = \frac{0}{14}. \tag{11}$$

These probability values can be used to generate new rule strings, or new solutions. Since the first rule in a solution has no parents, it will be chosen from nodes $N_{1j}$ according to their probabilities. The next rule will be chosen from nodes $N_{ij}$ according to the probabilities conditioned on the previous nodes. This building process is repeated until the last node has been chosen, i.e. node $N_{mj}$, where $m$ is number of the nurses. A link from nurse 1 to nurse $m$ is thus created, representing a new possible solution. Since all the probability values are normalized, we suggest the roulette-wheel method as suitable strategy for rule selection (Goldberg (1989)).

For further clarity, consider the following example of scheduling five nurses with two rules (1: random allocation, 2: allocate nurse to low-cost shifts). In the beginning of the search, the probabilities of choosing rule 1 or rule 2 for each nurse is equal, i.e. 50%. After a few iterations, due to the selection pressure and reinforcement learning, we experience two solution pathways: Because pure low-cost or random allocation produces low quality solutions, either rule 1 is used for the first 2-3 nurses and rule 2 on remainder or vice versa. In essence, the Bayesian network learns 'use rule 2 after 2-3x using rule 1' or vice versa.

**3.3 Our EDA approach**

Based on the estimation of conditional probabilities, this section introduces an EDA approach for the nurse scheduling problem. It uses techniques from the field of modelling data by Bayesian networks to estimate the joint distribution of promising solutions. The nodes, or variables, in the Bayesian network correspond to the individual rules by which a schedule will be built step by step.

In the proposed EDA, the first population of rule strings is generated at random. From the current population, a set of better rule strings is selected. Any selection method biased towards better fitness can be used, and in this paper, the traditional roulette-wheel selection is applied. The conditional probabilities of each node in the Bayesian network are computed. New rule strings are generated by using these conditional probability values, and are added into the old population, replacing some of the old rule strings. In more detail, the steps of the EDA for nurse scheduling are:

1. Set $t = 0$, and generate an initial population $P(0)$ at random;
2. Use roulette-wheel to select a set of promising rule strings $S(t)$ from $P(t)$ based on the rules in 4.4;
3. Compute the conditional probabilities of each node according to this set of promising solutions;
4. For the assignment of each nurse, the roulette-wheel method is used to select one rule according to the conditional probabilities of all available nodes, thus obtaining a new rule string. A set of new rule strings $O(t)$ will be generated in this way;
5. Create a new population $P(t+1)$ by replacing some rule strings from $P(t)$ with $O(t)$, and set $t = t+1$;
6. If the termination conditions are not met (we use 2000 generations), go to step 2.

**3.4 Four building rules**

Using our domain knowledge of nurse scheduling, the EDA can choose from the following four rules in this paper. Not that there are many more heuristic rules that could be used to build schedules.

3.4.1   Random rule

The first rule, called 'Random' rule, is used to select a nurse's shift pattern at random. Its purpose is to introduce randomness into the search thus enlarging the search space, and most importantly to ensure that the proposed algorithm has the ability to escape from local optima. This rule mirrors much of a scheduler's creativeness to come up with different solutions if required.

3.4.2   *k*-cheapest rule

The second rule is the '*k*-cheapest' rule. Disregarding the feasibility of the schedule, it randomly selects a shift pattern from a *k*-length list containing patterns with *k*-cheapest cost $p_{ij}$, in an effort to reduce the total cost of a schedule as much as possible.

3.4.3   Cover rule

Compared with the first two rules, the 'Cover' rule and fourth 'Contribution' rule are a little more complicated. The 'Cover' rule is designed to consider only the feasibility of the schedule. It schedules one nurse at a time in such a way as to cover those days and nights with the highest number of uncovered shifts.

The 'Cover' rule constructs solutions as follows. For each shift pattern in a nurse's feasible set, it calculates the total number of uncovered shifts that would be covered if the nurse worked that shift pattern. For instance, assume that a shift pattern covers Monday to Friday nights. Further assume that the current requirements for the nights from Monday to Sunday are as follows: (-3, 0, +1, -2, -1, -2, 0), where a negative number means undercover and a positive overcover. The Monday to Friday shift pattern hence has a cover value of 5. In this example, a Tuesday to Saturday pattern would have a value of 4.

In order to ensure that high-grade nurses are not 'wasted' covering unnecessarily lower grade shifts, for nurses of grade $s$, only shifts requiring grade $s$ nurses are counted as long as there is a single uncovered shift for this grade. If all these are covered, shifts of the next lower grade are considered and once these are filled those of the next lower grade. Due to the nature of this approach, nurses' preference costs $p_{ij}$ are not taken into account by this rule. However, they will influence decisions indirectly via the fitness function. Hence, the 'Cover' rule can be summarised as finding those shift patterns that cover the largest amount of undercover days.

3.4.4    Contribution rule

The fourth rule, called 'Contribution' rule, is biased towards solution quality but includes some aspects of feasibility by computing an overall score for each feasible pattern for the nurse currently being scheduled.

The 'Contribution' rule is designed to take into account the nurses' preferences. It therefore works with shift patterns rather than individual shifts. It also takes into account some of the covering constraints in which it gives preference to patterns that cover shifts that have not yet been allocated sufficient nurses to meet their total requirements. This is achieved by going through the entire set of feasible shift patterns for a nurse and assigning each one a score. The one with the highest (i.e. best) score is chosen. If there is more than one shift pattern with the best score, the first such shift pattern is chosen.

The score of a shift pattern is calculated as the weighted sum of the nurse's $p_{ij}$ value for that particular shift pattern and its contribution to the cover of all three grades. The latter is measured as a weighted sum of grade one, two and three uncovered shifts that would be covered if the nurse worked this shift pattern, i.e. the reduction in shortfall. Obviously, nurses can only contribute to uncovered demand of their own grade or below. More precisely and using the same notation as before, the score $p_{ij}$ of shift pattern $j$ for nurse $i$ is calculated with the following parameters:

- $d_{ks} = 1$ if there are still nurses needed on day $k$ of grade $s$ otherwise $d_{ks} = 0$;
- $a_{jk} = 1$ if shift pattern $j$ covers day $k$ otherwise $a_{jk} = 0$;
- $w_s$ is the weight of covering an uncovered shift of grade $s$;
- $w_p$ is the weight of the nurse's $p_{ij}$ value for the shift pattern.

Finally, $(100-p_{ij})$ must be used in the score, as higher $p_{ij}$ values are worse and the maximum for $p_{ij}$ is 100. Thus, the scores are calculated as follows:

$$S_{ij} = w_p(100 - P_{ij}) + \sum_{s=1}^{3} w_s q_{is} (\sum_{k=1}^{14} a_{jk} d_{ks}). \tag{12}$$

The 'Contribution' rule can be summarised as follows:

- Cycle through all shift patterns of a nurse;
- Assign each one a score based on covering uncovered shifts and preference cost;
- Choose the shift pattern with the highest score.

**3.5 Fitness function**

Independent of the rules used, the fitness of completed solutions has to be calculated. Unfortunately, feasibility cannot be guaranteed. This is a problem-specific issue and cannot be changed. Therefore, we still need a penalty function approach. Since the chosen encoding automatically satisfies constraint set (3) of the integer programming formulation, we can use the following formula, where $w_{demand}$ is the penalty weight, to calculate the fitness of solutions. Note that the penalty is proportional to the number of uncovered shifts.

$$\sum_{i=1}^{n}\sum_{j=1}^{m} p_{ij} x_{ij} + w_{demand} \sum_{k=1}^{14}\sum_{s=1}^{p} \max\left[R_{ks} - \sum_{i=1}^{n}\sum_{j=1}^{m} q_{is} a_{jk} x_{ij}; 0\right] \to \min!. \tag{13}$$

**4 Computational results**

**4.1 Details of algorithms**

To test the robustness of the proposed approach, each data set was run 20 times by fixing the parameters and varying the pseudo random number seeds. The results are listed in Table 1, in which N/A indicates no feasible solution was found and the last row contains the mean value of all columns. Note that when computing the mean, a censored cost value of 255 has been used in case an algorithm failed to find a feasible solution (N/A).

- IP: Optimal or best-known solutions found with IP software (Dowsland and Thompson (2000));
- GA: Best result out of 20 runs from a parallel genetic algorithm with multiple sub-populations and intelligent parameter adaptation (Aickelin and Dowsland (2000));
- Rd: the most simplified EDA, where only the random rule is used, i.e. equivalent to random search. It is worth noting that in this instance the Bayesian network reduces to one chain with a fixed structure and hence the algorithm becomes somewhat similar to mutual information maximisation for input clustering as presented by De Bonet et al (1997);
- CP: a simple EDA, where all four rules are used (see 4.4), but no conditional probabilities are computed, i.e. every rule has a 25% probability of being chosen all the time for all nurses;
- Op: Best result out of 20 runs of our standard EDA, i.e. four rules and conditional probabilities are used as described in section 3.1-3.4;
- Inf: Number of runs terminating with the best solution being infeasible;
- #: Number of runs terminating with the best solution being optimal or equal to the best known;
- <3: Number of runs terminating with the best solution being within three cost units of the optimum. The value of three units was chosen as it corresponds to the penalty cost of violating the least important level of requests in the original formulation. Thus, these solutions are still acceptable to the hospital.

For all data instances, the EDA used a set of fixed parameters as follows:

- Maximum number of generations = 2000;
- Penalty weight for each uncovered unit: $w_{demand}$ =200;
- For the '$k$-Cheapest' rule, $k$ = 5 (Aickelin and White (2004));
- Weight set for the 'Contribution' rule: w ={8,2,1,1};
- Population size = 140;
- Keep the best 40 solution in each generation;

- The executing time of the algorithm is approx. 10-20 seconds per run and data instance on a Pentium 4 PC.

### 4.2 Analysis of results

First, let us discuss the results in Table 1. Comparing the computational results on various test instances, one can see that using the random rule alone does not yield a single feasible solution. This underlines the difficulty of this problem. In addition, without learning the conditional probabilities, the results are much weaker, as the CP column shows. Thus, it is not simply enough to use the four rules to build solutions. Overall, the EDA's results found (Op) rival those found by the complex multi-population GA. For some data instances, the results are much better. Particular impressive is the fact that in 100% of cases a feasible solution is found. Note that independent of the algorithm used, some data instances are harder to solve than others due to a shortage of nurses in some weeks.

**Table 1 Comparison of results over 52 instances.**

Figures 4 and 5 show the results graphically. The bars above the *y*-axis represent solution quality. The black bars show the number of optimal (i.e. the value of '#' in Table 1), the grey near-optimal (i.e. the value of '<3' in Table 1) solutions. The bars below the y-axis represent the number of times the algorithm failed to find a feasible solution (i.e. the value of 'Inf' in Table 1). Hence, the shorter the bar is below the y-axis and the longer above, the better the algorithm's performance. Note that 'empty' bars mean that feasible, but not optimal solutions were found.

**Figure 4. Summary results of our EDA (Op).**

Figure 4 shows that for the EDA 38 out of 52 data sets are solved to or near to optimality. Additionally, feasible solutions are always found for all data sets and hence nothing is plotted below the x-axis.

For the GA (Aickelin and Dowsland (2003)) in figure 5 the results are similar: 42 data sets are solved well, however many solutions are infeasible and for two instances not a single feasible solution had been identified. Both algorithms have difficulties solving the later data sets (nurse shortages), but the EDA less so than the GA.

**Figure 5. Summary results of the adaptive multi-population GA.**

The behaviour of an individual run of the EDA is as expected. Figure 6 depicts the improvement of the schedule for the '04' data instance. At the $57^{th}$ generation, the optimal solution cost of 17 has been achieved. Although the actual values may differ among various instances, the characteristic shapes of the curves are similar for all seeds and data instances.

**Figure 6. Sample run of the EDA.**

To understand these results better, it is useful if we visualize the learning process in our EDA, rather than simply outputting the final computational results from a "black box". To achieve this target, we have developed a software package as a Java applet, which enables the algorithm to be run online via http://www.cs.nott.ac.uk/~jpl/BOA/BOA_Applet.html. Figure 7, 8 and 9 depict the graphic results of a single run (with a population of 100 and a specific random seed) in the beginning, intermediate and final stages of the evolutionary process respectively, for the '01' data instance which involves using four rules to schedule twenty-six nurses. In these graphs, the conditional probability of choosing a rule for a specific nurse has been mapped into a 256-gray value and each edge is denoted by drawing a line in this grey value. For instance, a probability of 50% would give a grey gradient of 128. The darker the edge, the higher the chance this edge (building block) will be used.

**Figure 7. Graphic demonstration of the probability distributions at the initial generation.**

**Figure 8. Graphic demonstration of the probability distributions after 50 generations.**

**Figure 9. Graphic demonstration of the probability distributions after 100 generations.**

The graphic results in Figure 7, 8 and 9 are in accordance with our hypothesis. In the early stages of the EDA, any edge (i.e. any building block) can be used to explore the solution space as much as possible, resulting in diversification. Hence, no particular edge stands out in Figure 7. With the EDA in progress, some edges are becoming clearer and clearer, while the ill-fitting paths are diminishing (Shown in Figure 8). Eventually, the algorithm converges to several optimum solutions with several clear paths, in which some edges might be common, such as the segment of path from node 10 to node 21 in Figure 9. This is particularly true for our particular problem instances because we know from the hospital that the optimum schedules are often not unique.

A direction for further research is to see if there are any good constructing sequences in individual data instances, for a fixed nurses' scheduling order. If so, the good patterns could be recognized and then extracted as new domain knowledge. Thus, by using the extracted knowledge, we can assign specific rules to the corresponding nurses beforehand, and only schedule the remaining nurses with all available rules, making it possible to reduce the solution space.

Figure 10 is the graphic demonstration of 20 runs of the proposed EDA for the '01' data instance. At each run, we evolved 100 populations and executed 100 generations with different random seeds. The results are interesting: no matter what parameter settings we have tried, nurse 9 is quite likely to be scheduled by rule 3, nurse 10 is the most likely to be scheduled by rule 4, and nurse 11 is quite likely to be scheduled by rule 3. Since this feature is very possible to appear in optimal solutions, it could be regarded as a good pattern.

**Figure 10. Summary results of 50 runs with different random seeds.**

The problem we are encountering is for most other data instances, good patterns are often not easy to be recognized by the method of observation. In this circumstance, the technology of pattern recognition might be applicable to extract elite patterns automatically in a statistical way.

## 5 Conclusions

A new scheduling algorithm based on Bayesian networks is presented in this paper. The approach is novel because it is the first time that Bayesian networks have been applied to the field of personnel scheduling. An effective method is proposed to solve the problem about how to implement explicit learning from past solutions. Unlike most existing rule-based approaches, the new approach has the ability to build schedules by using flexible, rather than fixed rules. Experimental results from real-world nurse scheduling problems have demonstrated the strength of the proposed approach.

Although we have presented this work in terms of nurse scheduling, it is suggested that the main idea of the approach could be applied to many other scheduling problems where the schedules will be built systematically according to specific rules. It is also hoped that this research will give some preliminary answers about how to include human-like learning into scheduling algorithms and may therefore be of interest to practitioners and researchers in areas of scheduling and evolutionary computation. In future, we will try to extract the 'explicit' part of the learning process further, e.g. by keeping partial solutions and learnt rules from one data instances to the next.

**Notes.**

## 1. Acknowledgements

The work was funded by the UK Government's major funding agency, Engineering and Physical Sciences Research Council (EPSRC), under grant GR/R92899/01.

## 2. Related information

JINGPENG LI*

jpl@cs.nott.ac.uk

School of Computer Science and IT, The University of Nottingham, Nottingham, NG8 1BB, United Kingdom

\* 1. Corresponding author.

   2. The authors are listed in alphabetical order.

# References.

**Tables.**

**Table 1. Comparison of results over 52 instances.**

| Set | IP | GA | Rd | CP | Op | Inf | # | <3 |
|---|---|---|---|---|---|---|---|---|
| *01* | 8 | 8 | N/A | 27 | 8 | 0 | 19 | 20 |
| *02* | 49 | 50 | N/A | 85 | 56 | 0 | 0 | 0 |
| *03* | 50 | 50 | N/A | 97 | 50 | 0 | 2 | 5 |
| *04* | 17 | 17 | N/A | 23 | 17 | 0 | 20 | 20 |
| *05* | 11 | 11 | N/A | 51 | 11 | 0 | 8 | 16 |
| *06* | 2 | 2 | N/A | 51 | 2 | 0 | 17 | 17 |
| *07* | 11 | 11 | N/A | 80 | 14 | 0 | 0 | 3 |
| *08* | 14 | 15 | N/A | 62 | 15 | 0 | 0 | 11 |
| *09* | 3 | 3 | N/A | 44 | 14 | 0 | 0 | 0 |
| *10* | 2 | 4 | N/A | 12 | 2 | 0 | 2 | 10 |
| *11* | 2 | 2 | N/A | 12 | 2 | 0 | 2 | 20 |
| *12* | 2 | 2 | N/A | 47 | 3 | 0 | 0 | 2 |
| *13* | 2 | 2 | N/A | 17 | 3 | 0 | 0 | 20 |
| *14* | 3 | 3 | N/A | 102 | 4 | 0 | 0 | 7 |
| *15* | 3 | 3 | N/A | 9 | 4 | 0 | 0 | 20 |
| *16* | 37 | 38 | N/A | 55 | 38 | 0 | 0 | 20 |
| *17* | 9 | 9 | N/A | 146 | 9 | 0 | 4 | 11 |
| *18* | 18 | 19 | N/A | 73 | 19 | 0 | 0 | 20 |
| *19* | 1 | 1 | N/A | 135 | 10 | 0 | 0 | 0 |
| *20* | 7 | 8 | N/A | 53 | 7 | 0 | 5 | 19 |
| *21* | 0 | 0 | N/A | 19 | 1 | 0 | 0 | 20 |
| *22* | 25 | 26 | N/A | 56 | 26 | 0 | 0 | 15 |
| *23* | 0 | 0 | N/A | 119 | 1 | 0 | 0 | 20 |
| *24* | 1 | 1 | N/A | 4 | 1 | 0 | 20 | 20 |
| *25* | 0 | 0 | N/A | 3 | 0 | 0 | 18 | 20 |
| *26* | 48 | 48 | N/A | 222 | 52 | 0 | 0 | 1 |
| *27* | 2 | 2 | N/A | 158 | 28 | 0 | 0 | 0 |
| *28* | 63 | 63 | N/A | 88 | 65 | 0 | 0 | 3 |
| *29* | 15 | 141 | N/A | 31 | 109 | 0 | 0 | 0 |
| *30* | 35 | 42 | N/A | 180 | 38 | 0 | 0 | 3 |
| *31* | 62 | 166 | N/A | 253 | 159 | 0 | 0 | 0 |
| *32* | 40 | 99 | N/A | 102 | 43 | 0 | 0 | 4 |
| *33* | 10 | 10 | N/A | 30 | 11 | 0 | 0 | 8 |
| *34* | 38 | 48 | N/A | 95 | 41 | 0 | 0 | 2 |
| *35* | 35 | 35 | N/A | 118 | 46 | 0 | 0 | 0 |
| *36* | 32 | 41 | N/A | 130 | 45 | 0 | 0 | 0 |
| *37* | 5 | 5 | N/A | 28 | 7 | 0 | 0 | 7 |
| *38* | 13 | 14 | N/A | 130 | 25 | 0 | 0 | 0 |
| *39* | 5 | 5 | N/A | 44 | 8 | 0 | 0 | 3 |
| *40* | 7 | 8 | N/A | 51 | 8 | 0 | 0 | 10 |
| *41* | 54 | 54 | N/A | 87 | 55 | 0 | 0 | 15 |
| *42* | 38 | 38 | N/A | 188 | 41 | 0 | 0 | 1 |
| *43* | 22 | 39 | N/A | 86 | 23 | 0 | 0 | 13 |
| *44* | 19 | 19 | N/A | 70 | 24 | 0 | 0 | 0 |
| *45* | 3 | 3 | N/A | 34 | 6 | 0 | 0 | 2 |
| *46* | 3 | 3 | N/A | 196 | 7 | 0 | 0 | 0 |
| *47* | 3 | 4 | N/A | 11 | 3 | 0 | 13 | 20 |
| *48* | 4 | 6 | N/A | 35 | 5 | 0 | 0 | 10 |
| *49* | 27 | 30 | N/A | 69 | 30 | 0 | 0 | 2 |
| *50* | 107 | 211 | N/A | 162 | 109 | 0 | 0 | 0 |
| *51* | 74 | N/A | N/A | 197 | 171 | 0 | 0 | 0 |
| *52* | 58 | N/A | N/A | 135 | 67 | 0 | 0 | 0 |
| *Av.* | **21** | **37** | **N/A** | **83** | **30** | **0** | **3** | **9** |

**Figures.**

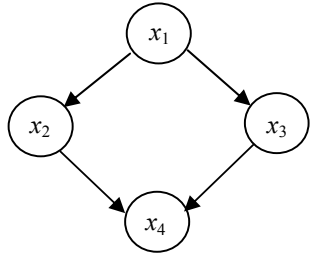

**Figure 1. Example of dependency graph with four nodes.**

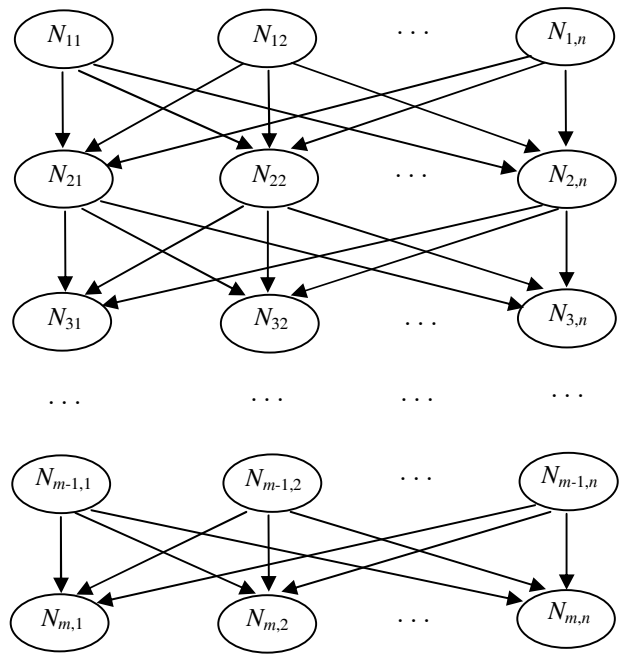

**Figure 2. A Bayesian network for nurse scheduling.**

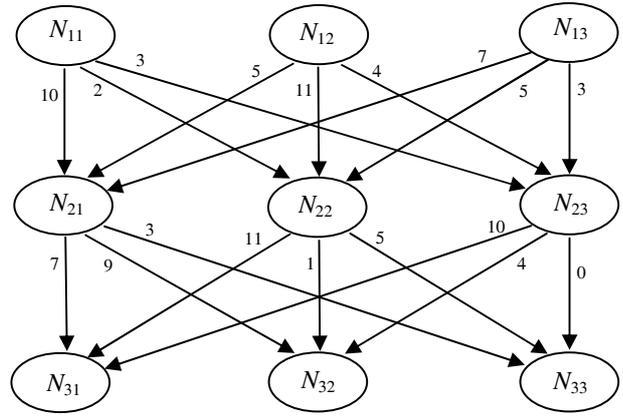

**Figure 3. A dummy problem with three nurses and three rules.**

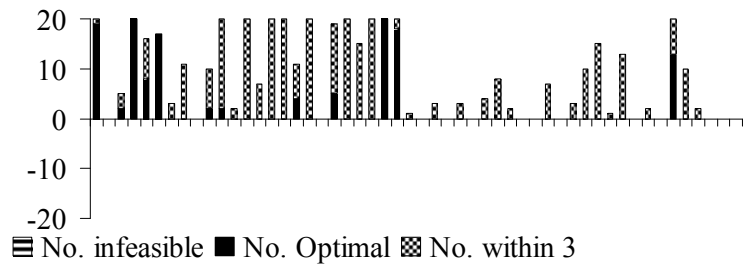

**Figure 4. Summary results of our EDA (Op).**

**GA**

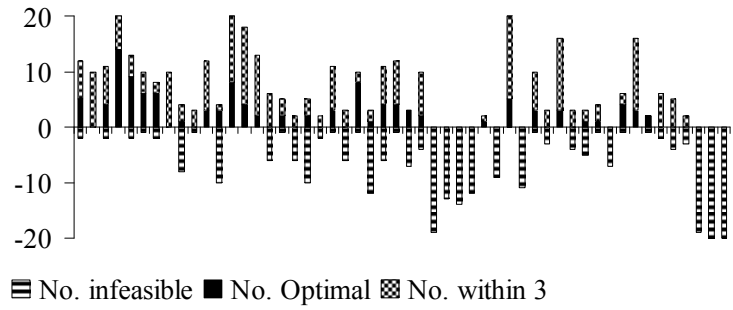

☰ No. infeasible ■ No. Optimal ▨ No. within 3

**Figure 5. Summary results of the adaptive multi-population GA.**

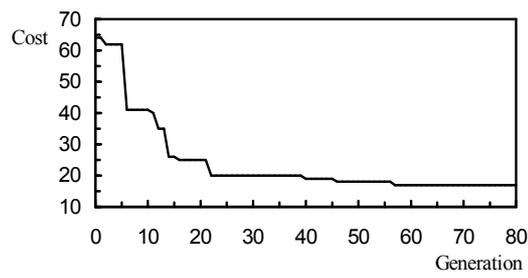

**Figure 6. Sample run of the EDA.**

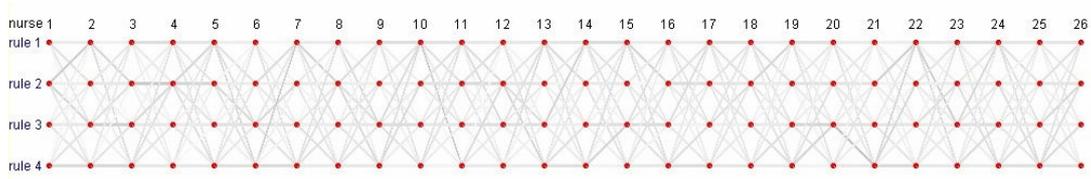

**Figure 7. Graphic demonstration of the probability distributions at the initial generation.**

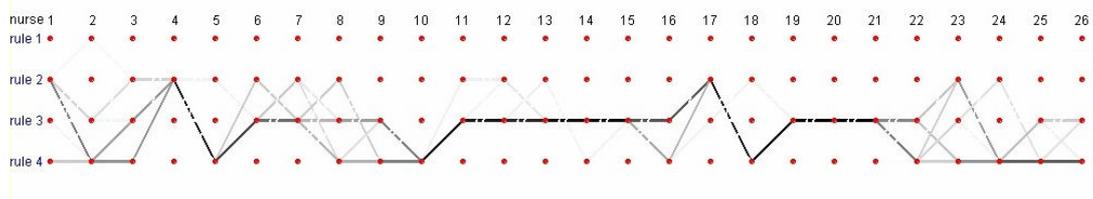

**Figure 8. Graphic demonstration of the probability distributions after 50 generations.**

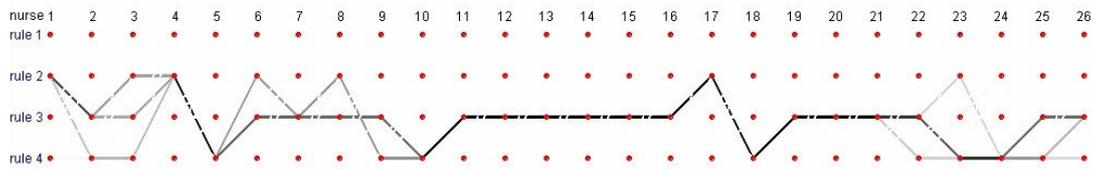

**Figure 9. Graphic demonstration of the probability distributions after 100 generations.**

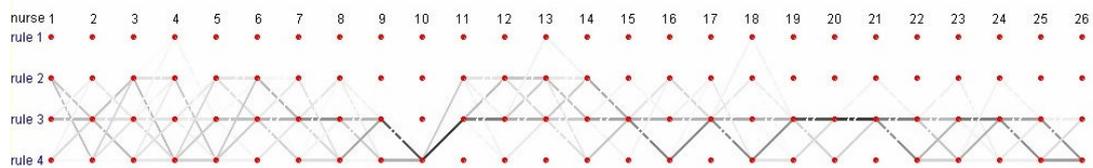

**Figure 10. Summary results of 50 runs with different random seeds.**